\documentclass[twoside,11pt]{article}
\usepackage{jmlr2e}

\usepackage[utf8]{inputenc}
\usepackage[T1]{fontenc}
\usepackage{url}
\usepackage{booktabs}
\usepackage{amsfonts}
\usepackage{amsmath}
\usepackage{amssymb}
\usepackage{nicefrac}
\usepackage[final]{microtype}
\usepackage{xcolor}
\usepackage{graphicx}
\usepackage{float}
\usepackage{algorithm}
\usepackage{algpseudocode}
\usepackage{multirow}
\usepackage{wrapfig}
\usepackage{subcaption}
\usepackage{enumitem}
\usepackage[capitalise]{cleveref}
\usepackage{pifont}

\newcommand{\cmark}{\ding{51}}
\newcommand{\xmark}{\ding{55}}

\newcommand{\method}{ConsistencyGate}

\newcommand{\eg}{\textit{e.g.,}}

\title{\method: Preventing Memory Contamination in LLM Agents\\
via Self-Consistency Admission Control}

\author{%
\name Yan Zhang \email yz18b@fsu.edu \\
\addr Department of Computer Science, Florida State University
\AND
\name Shibo Li\thanks{Corresponding author.} \email shiboli@cs.fsu.edu \\
\addr Department of Computer Science, Florida State University
}

\begin{document}
\firstpageno{1}

\begingroup
\renewcommand\thefootnote{\fnsymbol{footnote}}
\maketitle
\endgroup

\begin{abstract}
LLM agents that operate over many turns accumulate facts in an external memory
store and reuse them as premises for downstream reasoning.
A hallucinated fact written at one step therefore persists as a false premise for
every subsequent step, a failure mode we call \textbf{memory contamination}.
Existing memory management addresses retrieval and capacity but not write-time
correctness; this admission problem cannot be solved by utility- or recency-based
criteria, and uncontrolled contamination compounds across long trajectories.
We propose \textbf{\method}, a write-time admission gate that, before committing a
candidate fact $m$ extracted from context $c$, queries the LLM $K$ times for a
soft support score and admits $m$ only when the average exceeds a threshold $\tau$.
The mechanism is model-agnostic, requires no fine-tuning, and reduces to a single
forward pass in a log-probability variant for latency-sensitive deployments.
To measure the effect on natural data, we construct two real-conversation
benchmarks (\textsc{LoCoMo-Contam}, \textsc{MSC-Contam}) by planting controlled
single-detail corruptions in long-term conversations from LoCoMo and MSC, and
complement them with a structured synthetic corpus (\textsc{MemContam}) that
isolates a near-oracle upper bound.
Across four LLM backbones, \method{} reduces contamination on every benchmark
relative to a write-everything baseline, with the cost concentrated on facts that
are stated only implicitly in the source context.
We release all three benchmarks together with the gate implementation.
\end{abstract}

\section{Introduction}
\label{sec:intro}

Long-horizon LLM agents accumulate facts in an external memory store and reason over
them in later steps \citep{park2023generative, zhou2023agents, wang2023voyager, nakano2021webgpt}.
The architecture works only if the writer is reliable: a fact written once is read
back many times, so any error introduced at write time becomes a false premise for
every subsequent retrieval.
Yet LLMs hallucinate even when extracting from a given context
\citep{ji2023survey, maynez2020faithfulness, manakul2023selfcheckgpt}, so write-time errors
are not rare.
We study the resulting failure mode, which we call \textbf{memory contamination}:
the accumulation of hallucinated facts in agent memory, and the way these errors
compound over long interaction trajectories.

\begin{figure}[t]
    \centering
    \includegraphics[width=\linewidth]{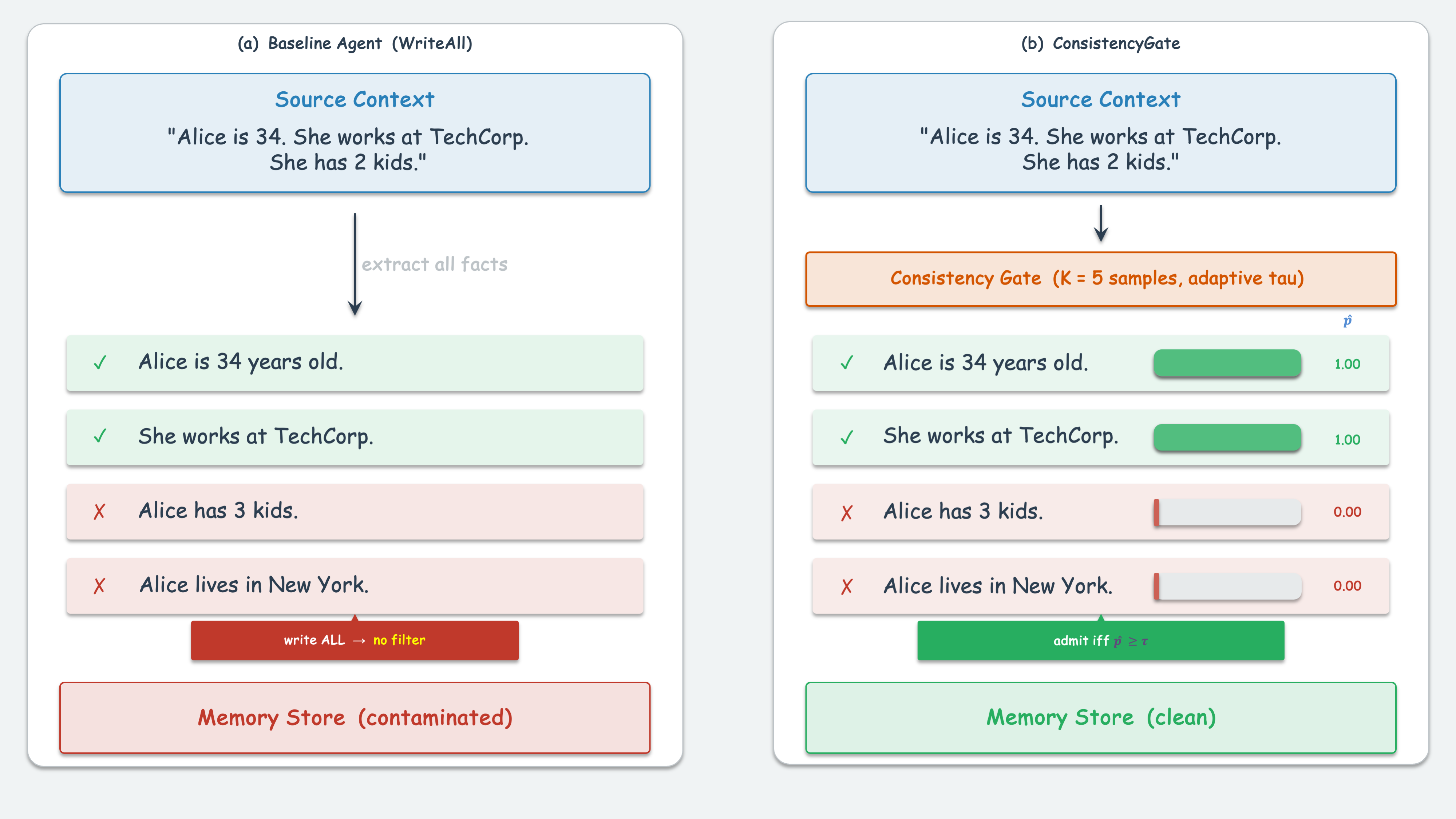}\\[0.4ex]
    \includegraphics[width=\linewidth]{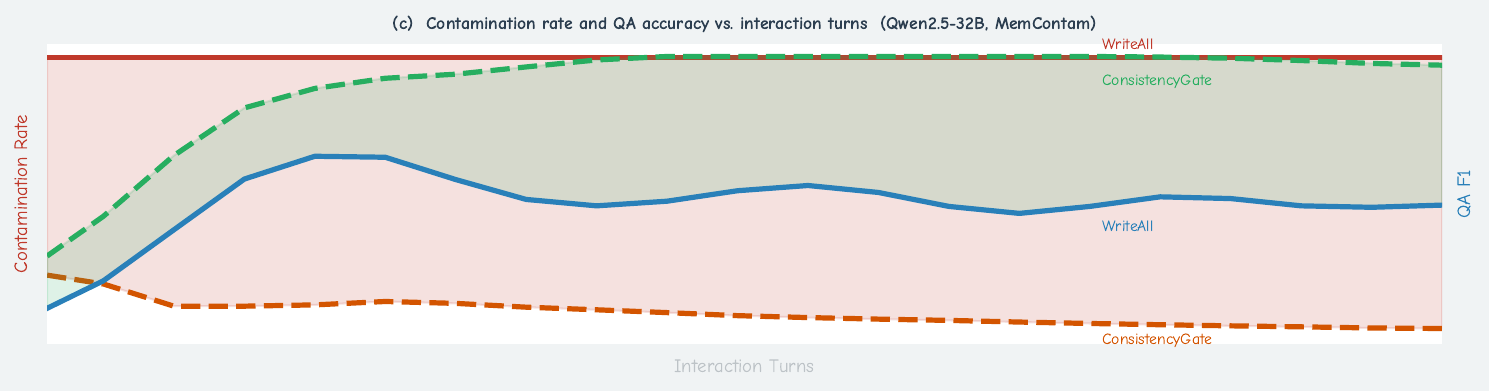}
    \caption{
        \textbf{Memory contamination and the \method{} mitigation.}
        \textit{(Left)} A baseline agent writes every extracted fact to memory,
        including hallucinated ones (red), which then propagate into downstream answers.
        \textit{(Right)} \method{} samples a verification score $K$ times before each
        write; facts with low average support are rejected and memory stays clean.
        \textit{(Bottom)} Contamination rate and downstream QA accuracy versus
        interaction turns on \textsc{MemContam} (Qwen2.5-32B-Instruct).
    }
    \label{fig:teaser}
\end{figure}

Memory contamination is distinct from the failure modes existing memory work addresses.
It is not a retrieval problem: once a fact is stored, even an oracle retriever will surface it.
It is not a capacity problem: contamination is not removed by eviction policies that
rank candidates on utility rather than correctness.
And it is not addressed by current admission policies, which score candidates on
novelty, recency, or future utility \citep{zhang2026amac}.
A hallucinated fact such as \textit{Alice graduated in 2015} (when context says 2012)
is as topically relevant, novel, and recent as the correct fact, so utility-based
admission is structurally blind to contamination.
Correctness is therefore an admission criterion that has no substitute in existing
memory machinery.

We propose \method, a write-time admission gate that scores candidate facts on
\emph{factual support} rather than utility.
Given a candidate fact $m$ extracted from context $c$, the gate queries the LLM $K$
times with a verification prompt and averages a soft 0--1 support score; $m$ is admitted
only when the average exceeds a threshold $\tau$.
The mechanism is model-agnostic, requires no fine-tuning, and reduces to a single
forward pass in a log-probability variant.
The threshold $\tau$ can be a fixed constant (we recommend $\tau{=}0.7$) or scheduled
adaptively against memory occupancy when per-deployment tuning is infeasible.

To measure contamination on natural data, we construct \textsc{LoCoMo-Contam} and
\textsc{MSC-Contam} from LoCoMo \citep{maharana2024locomo} and MSC \citep{xu2022msc}
by planting single-detail corruptions in probe facts.
We complement these with \textsc{MemContam}, a synthetic benchmark whose unambiguous
contexts isolate the gate's behavior in a structured setting.
Across the three benchmarks and four LLM backbones, \method{} consistently reduces
contamination, with the cost concentrated on facts that are stated only implicitly
in the source context.

\paragraph{Contributions.}
\textbf{(i)} We characterize \emph{memory contamination} as a write-time admission
problem distinct from retrieval and capacity, and show that it compounds across
interaction trajectories (\S\ref{sec:problem}).
\textbf{(ii)} We propose \method, a soft-scoring admission criterion that admits
a candidate fact only when its average support score across $K$ samples exceeds a
threshold $\tau$; the criterion has two efficiency modes ($K$-sample default and
a single-pass log-probability variant) and a deployment fallback (a
memory-occupancy-scaled threshold) when per-deployment tuning is infeasible
(\S\ref{sec:method}).
\textbf{(iii)} We construct three benchmarks: \textsc{MemContam} (synthetic),
\textsc{LoCoMo-Contam}, and \textsc{MSC-Contam}; to our knowledge the latter two are
the first real-conversation corpora with planted, controllable contamination
(\S\ref{sec:benchmark}). All three are released.
\textbf{(iv)} On the real-conversation benchmarks, \method{} reduces contamination
by 31.8\% and 26.6\% and lifts admission precision by 0.159 and 0.133 points on
Qwen2.5-32B, with consistent gains across three additional LLM backbones; on
\textsc{MemContam} it reaches a near-oracle 1.2\% contamination at 100\% recall
(\S\ref{sec:experiments}).

\section{Background and Related Work}
\label{sec:related}

\paragraph{Memory in LLM agents.}
External memory is a central component of long-horizon LLM agents
\citep{park2023generative, shinn2023reflexion, weng2023agent}, with episodic stores
for conversation history \citep{zhong2024memorybank, packer2023memgpt}, semantic
stores for extracted facts \citep{hu2023chatdb}, and procedural stores for learned
skills \citep{wang2023voyager}. We target the write process common to all three.
Prior memory management has focused on retrieval (which stored memories to surface
for a given query \citep{lewis2020rag, gao2023retrieval}) and capacity control
(what to evict when memory is full). The closest prior work is A-MAC \citep{zhang2026amac}, which formalizes admission
as a separate gate and scores candidate writes by a learned utility (novelty,
recency, projected retrieval value). A-MAC's objective is marginal usefulness;
\method{}'s is factual support against the source context, and the two are not
interchangeable: a correct fact and its hallucinated paraphrase have nearly
identical utility but opposite correctness scores. The two gates are therefore
complementary rather than competing (\S\ref{sec:contamination_analysis}).

\paragraph{Hallucination detection and correction.}
LLM hallucination is widely documented
\citep{ji2023survey, bang2023multitask, maynez2020faithfulness}, and a large
literature aims to detect or repair it.
Reference-free signals include atomic-fact decomposition with external evidence
(FactScore, \citealp{min2023factscore}), calibration via the model's own answer
probabilities \citep{kadavath2022know}, and self-feedback rewriting
(Self-Refine, \citealp{madaan2023refine}; Chain-of-Verification,
\citealp{dhuliawala2023cove}; RARR, \citealp{gao2023rarr}).
Retrieval-augmented variants such as Self-RAG \citep{asai2024selfrag}
and CRAG \citep{yan2024crag} insert validity checks between retrieval
and generation. \method{} sits at a different stage of the pipeline: rather than
verifying a generated answer or filtering retrieved evidence, it gates writes into
long-term memory before any downstream retrieval can amplify them. The closest
point of contact is SelfCheckGPT \citep{manakul2023selfcheckgpt}, which detects
hallucination in generated text by sampling multiple outputs and measuring
inter-sample agreement.

\paragraph{Self-consistency as a verification signal.}
\citet{wang2023selfconsistency} show that majority-vote aggregation over multiple
reasoning chains improves reasoning accuracy because correct answers are more
stable across samples than hallucinated ones. We adapt this principle to write-time
memory admission: the gate queries the LLM $K$ times and admits a candidate fact
only when the average support score exceeds a threshold. The key difference from
the prior self-consistency line is that the source context $c$ is available at
admission time, so the verifier grounds its judgment on a known evidence chunk
rather than on the model's prior alone.

\section{Problem Setup}
\label{sec:problem}

\subsection{Notation}

Let $\mathcal{M} = \{(m_i, t_i)\}$ denote an agent's memory store, where $m_i$ is a
text-form fact and $t_i$ is its timestamp.
At each step $t$, the agent observes context $c_t$ (a conversation turn, a document
chunk, or a tool output), and a \emph{memory writer} $\mathcal{W}$ extracts a set of
candidate facts $\mathcal{F}_t = \mathcal{W}(c_t)$.
A \emph{memory gate} $g: \mathcal{F}_t \times c_t \to \{0,1\}^{|\mathcal{F}_t|}$
decides which facts to admit; admitted facts are appended to $\mathcal{M}$.

We measure \textbf{memory contamination rate}:
\begin{equation}
    \rho \;=\; \frac{|\{m \in \mathcal{M} : \text{Verify}(m) = \text{False}\}|}{|\mathcal{M}|},
\end{equation}
where $\text{Verify}(m)$ returns the gold-label correctness of $m$ against its source
context. In our three benchmarks $\text{Verify}$ is provided by the construction
protocol (\S\ref{sec:benchmark}); in deployment it would be approximated by audit or
human review. \textbf{Admission precision} is $1 - \rho$; \textbf{admission recall}
is the fraction of correct candidates that the gate admits.

\subsection{Contamination Rates and Downstream Cost}
\label{sec:contamination_analysis}

Under the WriteAll baseline ($g \equiv 1$), $\rho = 50\%$ on all three contaminated
benchmarks (by construction), and downstream QA F1 collapses to 0.474 on
\textsc{MemContam}, 0.227 on \textsc{LoCoMo-Contam}, and 0.338 on \textsc{MSC-Contam}
(\Cref{tab:main_results}). The degradation is mechanistic: contaminated facts are
retrieved alongside correct ones and injected as evidence into the answer prompt,
so any retriever that does its job propagates the contamination.

\paragraph{Contamination cascades over a trajectory.}
Write-time contamination compounds with interaction length. Once a hallucinated fact
enters memory, an oracle retriever will still surface it; as memory fills, the
fraction of contaminated retrievals grows and downstream answers depend more on
false premises. We call this the \emph{contamination cascade} and quantify it in
\Cref{fig:cascade} (\S\ref{sec:experiments}): an admission error at turn $t$ continues
to corrupt turns $t{+}1, t{+}2, \ldots$ for the rest of the trajectory.

\paragraph{Utility-based admission cannot fix this.}
A hallucinated fact (\eg ``Alice graduated in 2015'' when context says 2012) is as
topically relevant, novel, and recent as the correct fact, so utility-based criteria
(relevance scoring, novelty filtering, recency management) are structurally blind
to contamination. The remedy is to score candidates on \emph{whether the source
context supports them}, which is the criterion \method{} operationalizes in
\S\ref{sec:method}.

\section{\method: Self-Consistency Memory Admission}
\label{sec:method}

\subsection{Core Algorithm}

\method{} wraps any existing memory writer with a consistency-checking gate.
Given a candidate fact $m$ extracted from context $c$, we score $m$'s reliability
by querying the LLM $K$ times with a \emph{soft} verification prompt that asks the
model to rate how strongly $m$ is supported by $c$ on a continuous 0--1 scale
(exact prompt in \Cref{app:prompts}):
\begin{equation}
    s_k = \mathcal{L}\bigl(\text{prompt}(m, c)\bigr) \in [0, 1], \quad k = 1, \ldots, K,
\end{equation}
where $s_k = 1$ means fully supported, $s_k = 0$ means contradicted, and
$s_k = 0.5$ means ambiguous.
The \textbf{consistency score} is the average support across $K$ samples,
\begin{equation}
    \hat{p}(m \mid c) = \frac{1}{K} \sum_{k=1}^{K} s_k,
\end{equation}
and the admission decision is $g(m, c) = \mathbf{1}[\hat{p}(m \mid c) \geq \tau]$.
\Cref{alg:consistency_gate} gives the full procedure. We recommend the fixed
setting $\tau{=}0.7$, $K{=}5$ as a default (see \S\ref{sec:ablation}), with the
adaptive schedule of \S\ref{sec:adaptive} as a zero-tuning fallback.

\begin{algorithm}[t]
\caption{\method{} Memory Admission}
\label{alg:consistency_gate}
\begin{algorithmic}[1]
\Require Candidate fact $m$, source context $c$, LLM $\mathcal{L}$,
         samples $K$, threshold $\tau$, memory $\mathcal{M}$, capacity $C$
\State $\text{total} \leftarrow 0$
\For{$k = 1$ to $K$}
    \State $\text{prompt} \leftarrow$
        \textsc{SoftVerifyPrompt}$(m, c)$
    \State $s_k \leftarrow$
        $\mathcal{L}(\text{prompt}) \in [0, 1]$
    \State $\text{total} \leftarrow \text{total} + s_k$
\EndFor
\State $\hat{p} \leftarrow$
    \text{total} / K
\State $\tau^* \leftarrow \tau$
    \Comment{or \textsc{AdaptiveThreshold}$(\tau_{\min}, \tau_{\max}, |\mathcal{M}|, C)$;
             see Eq.~\ref{eq:adaptive}}
\If{$\hat{p} \geq \tau^*$}
    \State $\mathcal{M}.\text{add}(m)$
    \State \Return \textbf{True}
\EndIf
\State \Return \textbf{False}
\end{algorithmic}
\end{algorithm}

\subsection{Efficient Single-Pass Approximation}
\label{sec:logprob_method}

The default soft-scoring implementation issues $K$ LLM calls per candidate fact,
which becomes expensive when the writer extracts many facts per context.
We provide an alternative that uses \textbf{token-level log-probabilities}: switch
to a binary yes/no prompt and read the score from a single forward pass,
\begin{equation}
    \hat{p}_{\log}(m \mid c) = \mathrm{softmax}\!\left(\bigl[\log p(\texttt{yes} \mid \text{prompt}), \; \log p(\texttt{no} \mid \text{prompt})\bigr]\right)\!\bigl[0\bigr],
    \label{eq:logprob}
\end{equation}
which is exposed by most LLM serving stacks.
The variant runs in 23--28\,ms per fact across our four backbones (versus
264--363\,ms at $K{=}5$, a $12{-}14\times$ speedup) and matches or beats soft
scoring on structured contexts.
We name this asymmetry the \emph{bimodality of yes/no support}: on structured
contexts the model's yes/no token probabilities saturate
($\log P(\text{yes} \mid c, m_+) \to 0$ for correct $m_+$,
$\log P(\text{no} \mid c, m_-) \to 0$ for corrupted $m_-$), so $\hat{p}_{\log}$
approximates $\hat{p}$ in a single forward pass; on long-form conversational
contexts the bimodal assumption breaks down and $K$-sample averaging recovers an
estimate that one logprob query cannot (\Cref{sec:logprob_analysis}).

\subsection{Adaptive Thresholding}
\label{sec:adaptive}

When $\tau$ cannot be tuned for a particular model or deployment, we provide an
\textbf{adaptive schedule} that scales the threshold with memory utilization:
\begin{equation}
    \tau^*(|\mathcal{M}|, C; \tau_{\min}, \tau_{\max}) =
    \tau_{\min} + (\tau_{\max} - \tau_{\min}) \cdot \frac{|\mathcal{M}|}{C},
    \label{eq:adaptive}
\end{equation}
where $C$ is the memory capacity, $\tau_{\min}$ is the floor at empty memory, and
$\tau_{\max}$ the ceiling at full capacity.
The schedule is monotone in occupancy: more permissive when memory is sparse, and
stricter as memory fills.
When the operating point can be tuned for the deployment, fixed $\tau{=}0.7$
outperforms the adaptive schedule (\S\ref{sec:ablation}); the adaptive variant is
recommended only when per-deployment tuning is infeasible.

\section{Benchmarks}
\label{sec:benchmark}

Existing memory benchmarks \citep{maharana2024locomo, xu2022msc} lack gold labels for
fact-level accuracy, making it difficult to measure contamination rates precisely.
We introduce three complementary benchmarks: a controlled synthetic corpus
(\textsc{MemContam}) and two real-conversation corpora derived from existing
long-term dialogue datasets (\textsc{LoCoMo-Contam}, \textsc{MSC-Contam}).
All three impose a controlled 50\% contamination rate so that the contamination
metric is comparable across them.

\paragraph{Evaluation metrics (shared across benchmarks).}
We report:
(1)~\textbf{Contamination rate} $\rho$ (\% incorrect among admitted facts, lower is
better);
(2)~\textbf{Admission precision} (\% correct among admitted; equal to $1-\rho$);
(3)~\textbf{Admission recall} (\% of correct facts admitted); and
(4)~\textbf{Downstream QA F1} on question-answer pairs answerable from the correct
facts.

\subsection{\textsc{MemContam} (Synthetic)}
\label{sec:memcontam}

\paragraph{Construction.}
We define five fixed source contexts spanning diverse domains (personal biography,
library record, recipe, science fair report, spacecraft mission log).
Each context is used to generate 40 independent samples via Qwen2.5-32B-Instruct,
yielding 200 samples in total with broad lexical variation despite shared
underlying entities.
For each sample, the LLM extracts $F=5$ ground-truth facts and generates $F'=5$
plausible but incorrect variants using four corruption strategies:
(a)~numerical perturbation, (b)~negation, (c)~proper-noun substitution, and
(d)~appended contradicting clause.
The benchmark contains 1{,}000 correct and 1{,}000 incorrect facts across the
200 contexts, yielding the controlled 50\% contamination setting.
Because corruption (d) appends a literal contradicting clause and the verifier
shares a model family with the corpus generator, we treat \textsc{MemContam}
results as an \emph{internal-consistency upper bound} rather than a deployment
headline; deployment-relevant numbers come from \textsc{LoCoMo-Contam} and
\textsc{MSC-Contam}.

\subsection{\textsc{LoCoMo-Contam} (Real Long-Term Conversations)}
\label{sec:locomo_contam}

\paragraph{Source data.}
\textsc{LoCoMo-Contam} is built on top of LoCoMo-10
\citep{maharana2024locomo}, a set of 10 long-term dyadic conversations
(400--700 turns each, 19--32 sessions per conversation).

\paragraph{Construction.}
For each of the 10 conversations we select up to $N{=}5$ QA pairs whose answers are
concrete and verifiable (factual or temporal categories).
Each QA pair is converted into a self-contained fact statement using
Qwen2.5-32B-Instruct (\eg ``Caroline moved from Sweden 4 years ago.''), and a
\emph{corrupted} version is generated by prompting the model to change exactly one
concrete detail to a plausible but incorrect alternative
(\eg ``Caroline moved from \textbf{Germany} 4 years ago.'').
Both versions are then paraphrased into natural conversational turns as if spoken
by one of the participants.
The resulting benchmark contains 50 probe pairs (5 per conversation) at 50\%
contamination.

\paragraph{Evaluation protocol.}
The agent first processes the full conversation to populate memory.
Correct probe facts are then pre-seeded directly into memory (bypassing the gate)
to ensure that retrieval evidence is always available; this design choice means
downstream QA is insensitive to gate decisions on probe turns, so the primary
signals on this benchmark are the contamination rate and admission precision
(\S\ref{sec:experiments}).
Each probe turn (correct and corrupted) is then presented through the full
observe pipeline: the memory writer extracts a fact, the gate queries memory for
supporting context, and an admission decision is recorded.
This tests whether the gate can distinguish a claim consistent with established
memory from one that contradicts it, on real conversational data.

\subsection{\textsc{MSC-Contam} (Persona-Grounded Conversations)}
\label{sec:msc_contam}

\paragraph{Source data.}
\textsc{MSC-Contam} is built on top of the Multi-Session Chat dataset
\citep{xu2022msc}, a corpus of dyadic conversations grounded in evolving speaker
personas (20 conversations, 2 sessions each).

\paragraph{Construction.}
Unlike LoCoMo, MSC does not come with pre-annotated QA pairs; each speaker is
instead seeded with a set of persona sentences at the start of each session.
We filter persona sentences for concreteness (\eg ``I have worked as a nurse for
10 years'' vs.\ ``I enjoy being outdoors''), keeping only those with a verifiable
concrete detail (name, number, place, or occupation), and select up to $N{=}5$
persona facts per conversation.
For each fact, we generate a corrupted version by changing exactly one concrete
detail to a plausible but incorrect alternative, using the same corruption prompts
as \textsc{LoCoMo-Contam}.
Both versions are then converted into natural conversational utterances, and QA
pairs for downstream evaluation are generated directly from the original persona
facts.
The resulting benchmark contains 41 probe pairs (up to 5 per conversation
$\times$ 20 conversations; some conversations contain fewer than 5 concrete
persona facts) at the same 50\% contamination rate.

\paragraph{Evaluation protocol.}
Identical to \textsc{LoCoMo-Contam}: the agent processes all session turns to
populate memory, correct facts are pre-seeded, and each probe turn (correct and
corrupted) is evaluated through the full gate pipeline.
\textsc{MSC-Contam} differs from \textsc{LoCoMo-Contam} in that its sessions are
shorter and more topically focused, and probe facts are persona-grounded rather
than scattered across many conversational turns; this lets us test whether the
gate can leverage persona memory established across multiple sessions.

\section{Experiments}
\label{sec:experiments}

\subsection{Experimental Setup}

All agents use \texttt{Qwen2.5-32B-Instruct} (bfloat16, served via vLLM~\citep{kwon2023vllm} on a single high-memory GPU sufficient for 32B-parameter serving) with sentence-transformer retrieval over a flat memory store; the memory writer extracts atomic facts in one pass per context chunk. \textsc{MemContam} evaluates the gate directly on gold candidate facts (bypassing the extraction step) to isolate admission accuracy; \textsc{LoCoMo-Contam} and \textsc{MSC-Contam} run the full agent pipeline. We compare two baselines: \textbf{WriteAll} (admits every fact) and \textbf{Random} (admits each fact with probability $p{=}0.6$, matched to \method{}'s empirical admission rate at $\tau{=}0.7$ to isolate the contribution of correctness-based filtering from any reduction in memory size). \textbf{\method{}} uses our recommended configuration of fixed threshold $\tau{=}0.7$ with $K{=}5$ samples and capacity $C{=}500$ (chosen larger than the maximum candidate count per trajectory so capacity does not bind in any reported run); the adaptive threshold schedule and the single-pass LogProb variant are evaluated separately in \S\ref{sec:ablation}. Cross-model experiments on Llama-3.3-70B and Llama4-Scout-17B-16E use two high-memory GPUs in tensor-parallel mode.

\paragraph{LLM use disclosure.}
The LLM plays three roles: verifier inside \method{}, memory writer (candidate extraction), and corpus/corruption generator for benchmark construction (\S\ref{sec:benchmark}); the verifier and writer use whichever backbone is under evaluation, while the generator uses Qwen2.5-32B-Instruct only.

\subsection{Main Results}

\begin{table}[t]
\centering
\caption{
    Main results across four benchmarks; \method{} uses our recommended
    configuration (fixed $\tau{=}0.7$, $K{=}5$).
    \textbf{LoCoMo}: clean, tests over-rejection.
    \textbf{LoCoMo-Contam}/\textbf{MSC-Contam}: real conversations with planted
    corruptions; $\rho$ and Prec are the primary metrics (QA F1 is insensitive due to
    pre-seeding, see \S\ref{sec:locomo_contam}).
    \textbf{\textsc{MemContam}}: synthetic, controlled 50\% contamination.
    $\rho$: contamination rate (\%$\downarrow$). Prec/Rec: admission precision/recall
    ($\uparrow$). F1: downstream QA ($\uparrow$). Rec of WriteAll is trivially 1.0.
    \textbf{Bold}: best per column.
}
\label{tab:main_results}
\resizebox{\linewidth}{!}{%
\begin{tabular}{lccccccccccccc}
\toprule
& \textbf{LoCoMo} & \multicolumn{4}{c}{\textbf{LoCoMo-Contam}} & \multicolumn{4}{c}{\textbf{MSC-Contam}} & \multicolumn{4}{c}{\textbf{\textsc{MemContam}}} \\
\cmidrule(lr){2-2}\cmidrule(lr){3-6}\cmidrule(lr){7-10}\cmidrule(lr){11-14}
\textbf{Method} & QA F1$\uparrow$
& $\rho\downarrow$ & Prec$\uparrow$ & Rec$\uparrow$ & QA F1$\uparrow$
& $\rho\downarrow$ & Prec$\uparrow$ & Rec$\uparrow$ & QA F1$\uparrow$
& $\rho\downarrow$ & Prec$\uparrow$ & Rec$\uparrow$ & QA F1$\uparrow$ \\
\midrule
WriteAll           & \textbf{0.271} & 50.0 & 0.500 & \textbf{1.00} & 0.227 & 50.0 & 0.500 & \textbf{1.00} & 0.338 & 50.0 & 0.500 & \textbf{1.00} & 0.474 \\
Random             & 0.227 & 47.7 & 0.523 & 0.90 & 0.206 & 49.3 & 0.507 & 0.83 & \textbf{0.357} & 49.3 & 0.507 & 0.60 & 0.443 \\
\midrule
\textbf{\method{}} & 0.267 & \textbf{34.1} & \textbf{0.659} & 0.58 & \textbf{0.230}
  & \textbf{36.7} & \textbf{0.633} & 0.93 & 0.356
  & \textbf{1.2} & \textbf{0.988} & \textbf{1.00} & \textbf{0.840} \\
\bottomrule
\end{tabular}%
}
\end{table}

\Cref{tab:main_results} shows our main results across all four benchmarks (\textsc{MemContam} over 3 seeds).

\paragraph{\textsc{MemContam}.}
On the synthetic benchmark, \method{} reduces contamination from 50.0\% to 1.2\%
(a 97.6\% relative reduction) at 100\% recall of correct facts, and lifts downstream
QA F1 from 0.474 to 0.840.
We read this number as a near-oracle upper bound rather than a deployment number:
\textsc{MemContam} contexts are short, factually unambiguous, and the corruption
strategies leave differences large enough that a same-family LLM verifier resolves
them with very few samples (see \S\ref{sec:logprob_analysis}).
The Random baseline does not improve over WriteAll, ruling out the explanation that
any selective admission helps.

\paragraph{\textsc{LoCoMo-Contam} and \textsc{MSC-Contam}.}
On real long-term conversations with planted corruptions, \method{} reduces
contamination from 50.0\% to 34.1\% on \textsc{LoCoMo-Contam} (precision
$0.500{\to}0.659$) and to 36.7\% on \textsc{MSC-Contam} (precision
$0.500{\to}0.633$); QA F1 is nearly unchanged because pre-seeding correct facts
makes downstream QA insensitive to gate decisions on probe turns
(\S\ref{sec:locomo_contam}).
Recall on \textsc{LoCoMo-Contam} is 0.58 (systematic across all four backbones,
range 0.52--0.68; \Cref{sec:cross_model}) versus 0.93 on \textsc{MSC-Contam}:
persona facts are stated explicitly and self-contained, whereas LoCoMo facts are
often implicit or distributed across many turns and the verifier prompt requires
support to be \emph{explicit}. The deployment-side trade-off is discussed in
\S\ref{sec:analysis} and \S\ref{sec:limitations}.

\paragraph{LoCoMo (clean).}
On clean conversations \method{} reaches QA F1 $=0.267$ versus 0.271 for
WriteAll ($-1.5\%$); Random filtering at a comparable admission rate is
substantially worse (F1 $=0.227$), indicating that the gain on contaminated data
comes from correctness-based filtering rather than from any reduction in memory
size.

\subsection{Cross-Model Generalization}
\label{sec:cross_model}

\begin{table}[t]
\centering
\caption{
    Cross-model generalization of \method{} ($K{=}5$). Qwen2.5-32B and Llama-3.3-70B
    use fixed $\tau{=}0.7$; Gemma-31B and Llama4-Scout report the adaptive schedule
    ($\tau_{\min}{=}0.4, \tau_{\max}{=}0.8$) used in their respective cross-model
    runs.
    \method{} QA F1 on \textsc{MemContam} is the mean over 3 seeds.
    $\rho$: contamination rate (\%). Prec/Rec: admission precision/recall.
    Rec of WriteAll is trivially 1.0. \textbf{Bold}: best per model group.
}
\label{tab:cross_model}
\resizebox{\linewidth}{!}{%
\begin{tabular}{llccccccccccccc}
\toprule
& & \textbf{LoCoMo}
& \multicolumn{4}{c}{\textbf{LoCoMo-Contam}}
& \multicolumn{4}{c}{\textbf{MSC-Contam}}
& \multicolumn{4}{c}{\textbf{\textsc{MemContam}}} \\
\cmidrule(lr){3-3}\cmidrule(lr){4-7}\cmidrule(lr){8-11}\cmidrule(lr){12-15}
\textbf{Model} & \textbf{Method}
& QA F1$\uparrow$
& $\rho\downarrow$ & Prec$\uparrow$ & Rec$\uparrow$ & QA F1$\uparrow$
& $\rho\downarrow$ & Prec$\uparrow$ & Rec$\uparrow$ & QA F1$\uparrow$
& $\rho\downarrow$ & Prec$\uparrow$ & Rec$\uparrow$ & QA F1$\uparrow$ \\
\midrule
\multirow{2}{*}{Qwen2.5-32B}
  & WriteAll  & \textbf{0.271} & 50.0 & 0.500 & \textbf{1.00} & 0.227 & 50.0 & 0.500 & \textbf{1.00} & 0.338 & 50.0 & 0.500 & \textbf{1.00} & 0.474 \\
  & \method{} & 0.267 & \textbf{34.1} & \textbf{0.659} & 0.58 & \textbf{0.230} & \textbf{36.7} & \textbf{0.633} & 0.93 & \textbf{0.356} & \textbf{1.2} & \textbf{0.988} & \textbf{1.00} & \textbf{0.840} \\
\midrule
\multirow{2}{*}{Gemma4-31B}
  & WriteAll  & 0.265 & 50.0 & 0.500 & \textbf{1.00} & 0.307 & 50.0 & 0.500 & \textbf{1.00} & 0.398 & 50.0 & 0.500 & \textbf{1.00} & 0.471 \\
  & \method{} & \textbf{0.269} & \textbf{18.9} & \textbf{0.811} & 0.60 & \textbf{0.316} & \textbf{26.9} & \textbf{0.731} & 0.93 & \textbf{0.407} & \textbf{3.4} & \textbf{0.966} & \textbf{1.00} & \textbf{0.831} \\
\midrule
\multirow{2}{*}{Llama4-Scout-17B}
  & WriteAll  & \textbf{0.239} & 50.0 & 0.500 & \textbf{1.00} & 0.194 & 50.0 & 0.500 & \textbf{1.00} & 0.441 & 50.0 & 0.500 & \textbf{1.00} & 0.448 \\
  & \method{} & 0.202 & \textbf{29.7} & \textbf{0.703} & 0.52 & \textbf{0.208} & \textbf{40.0} & \textbf{0.600} & 0.95 & \textbf{0.452} & \textbf{14.0} & \textbf{0.860} & \textbf{1.00} & \textbf{0.693} \\
\midrule
\multirow{2}{*}{Llama-3.3-70B}
  & WriteAll  & \textbf{0.285} & 50.0 & 0.500 & \textbf{1.00} & \textbf{0.286} & 50.0 & 0.500 & \textbf{1.00} & 0.443 & 50.0 & 0.500 & \textbf{1.00} & 0.433 \\
  & \method{} & 0.265 & \textbf{19.0} & \textbf{0.810} & 0.68 & 0.263 & \textbf{32.1} & \textbf{0.679} & 0.93 & \textbf{0.448} & \textbf{0.2} & \textbf{0.998} & \textbf{1.00} & \textbf{0.819} \\
\bottomrule
\end{tabular}%
}
\end{table}

\Cref{tab:cross_model} reports \method{} on three additional backbones.
On \textsc{MemContam} all four models reduce contamination to single-digit
percentages and lift QA F1 by 75--89\% relative to WriteAll; on real conversations
the reductions span 20\% (Scout, \textsc{MSC-Contam}) to 62\% (Gemma and
Llama-3.3-70B, \textsc{LoCoMo-Contam}), with clean-LoCoMo regression within four
points throughout.
Llama-3.3-70B is the most reliable verifier (precision $=0.998$ across every
$K{\in}\{1,3,5,10\}$ and $\tau{\in}\{0.3,0.5,0.7,0.9\}$, plus the best
precision/recall balance on \textsc{LoCoMo-Contam} at Prec $=0.810$,
Recall $=0.680$); Llama4-Scout is the weakest, with its broader soft-score
distribution admitting borderline facts at low thresholds ($\rho{=}14.0\%$ under
the adaptive schedule used here, dropping to $0.30\%$ at $\tau{=}0.7$;
\Cref{sec:ablation}).
Verifier capability matters but is not a binary requirement: every backbone we
tested reduces contamination, and the recommended setting closes most of the
cross-model gap.

\subsection{Contamination Cascade Effect}

\begin{figure}[t]
    \centering
    \includegraphics[width=0.55\linewidth]{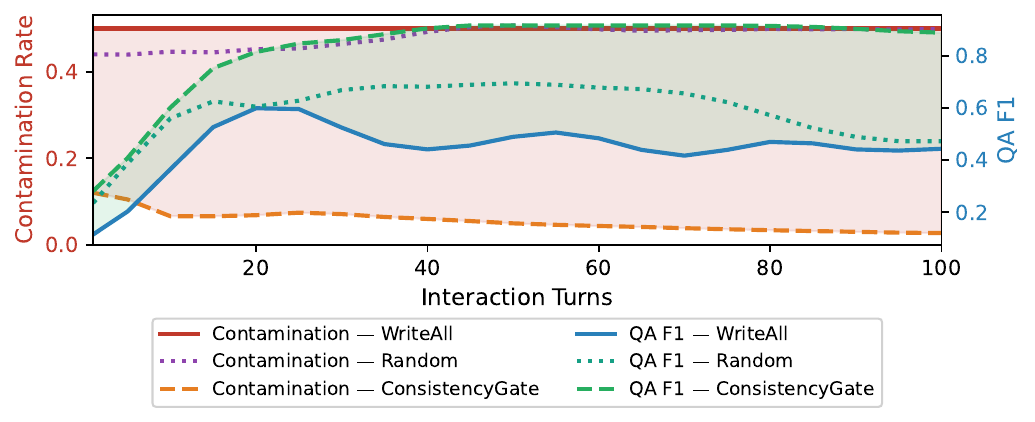}
    \caption{Contamination rate and QA F1 vs.\ interaction turns on \textsc{MemContam} (Qwen2.5-32B-Instruct).
    Under WriteAll, contamination stays at 50\% throughout and QA F1 saturates at 0.45 as incorrect facts crowd out correct ones.
    \method{} reduces contamination steadily, falling below 3\% by turn 90, and QA F1 rises to 0.88 as clean facts accumulate.}
    \label{fig:cascade}
\end{figure}

\Cref{fig:cascade} traces contamination rate and QA F1 across 100 interaction turns on
\textsc{MemContam}.
Under WriteAll, contamination stays at 50\% and QA F1 saturates at 0.45 as incorrect
facts crowd out correct ones in retrieval.
Under \method{}, contamination falls below 3\% by turn 90 and QA F1 climbs to 0.88;
the gap between the two methods widens monotonically with trajectory length.
\textsc{MemContam}'s controlled setting makes this trajectory-level measurement possible;
replicating it on real conversations requires turn-level fact labels that current corpora
do not provide at scale.

\subsection{Ablation Studies}
\label{sec:ablation}

\begin{table}[t]
\centering
\caption{Ablation summary on \textsc{MemContam}: key variants per model.
Full results ($K \in \{1,3,5,10\}$, all $\tau$ values) in \cref{tab:ablation_full}.
\textbf{Bold}: best per model group.}
\label{tab:ablation}
\small
\setlength{\tabcolsep}{4pt}
\begin{tabular}{llcccc}
\toprule
& \textbf{Variant} & \textbf{Adm Prec} & \textbf{ContamRate} & \textbf{QA F1} & \textbf{Lat (ms)} \\
\midrule
\multicolumn{6}{@{}l}{\textit{Qwen2.5-32B-Instruct}} \\[0.3ex]
& \method{} ($K=5$, default) & 0.941 & 5.93\% & 0.803 & 341.9 \\
& Fixed $\tau = 0.7$         & 0.988 & 1.19\% & \textbf{0.842} & 341.6 \\
& LogProb                    & \textbf{1.000} & \textbf{0.00\%} & 0.824 & \textbf{27.5} \\
\midrule
\multicolumn{6}{@{}l}{\textit{Gemma4-31B-IT}} \\[0.3ex]
& \method{} ($K=5$, default) & 0.966 & 3.38\% & 0.833 & 339.6 \\
& Fixed $\tau = 0.7$         & \textbf{0.984} & \textbf{1.57\%} & \textbf{0.847} & 339.7 \\
& LogProb                    & \textbf{0.984} & \textbf{1.57\%} & \textbf{0.847} & \textbf{24.3} \\
\midrule
\multicolumn{6}{@{}l}{\textit{Llama4-Scout-17B-16E}} \\[0.3ex]
& \method{} ($K=5$, default) & 0.861 & 13.9\% & 0.716 & 264.6 \\
& Fixed $\tau = 0.7$         & 0.997 & 0.30\% & 0.783 & 260.3 \\
& LogProb                    & \textbf{0.999} & \textbf{0.10\%} & \textbf{0.784} & \textbf{23.1} \\
\midrule
\multicolumn{6}{@{}l}{\textit{Llama-3.3-70B-Instruct}} \\[0.3ex]
& \method{} ($K=5$, default) & 0.998 & 0.20\% & 0.818 & 363.4 \\
& Fixed $\tau = 0.7$         & 0.998 & 0.20\% & 0.814 & 363.0 \\
& LogProb                    & \textbf{1.000} & \textbf{0.00\%} & \textbf{0.819} & \textbf{25.7} \\
\bottomrule
\end{tabular}
\end{table}

Increasing $K$ produces only marginal precision gains (1--2 points on Qwen and
Gemma between $K{=}1$ and $K{=}10$, essentially flat on Scout and
Llama-3.3-70B), so we recommend $K{=}5$ as a conservative default
($\leq{}365$\,ms per fact) and $K{=}1$ for latency-critical settings.
Raising $\tau$ from 0.3 to 0.7 sharply improves precision on the three more
sensitive backbones (Qwen $0.931{\to}0.988$, Gemma $0.966{\to}0.984$,
Scout $0.860{\to}0.997$), and $\tau{=}0.9$ matches $\tau{=}0.7$ within noise,
identifying $\tau{=}0.7$ as the natural decision boundary; the adaptive schedule
($\tau_{\min}{=}0.4, \tau_{\max}{=}0.8$) is a no-tuning fallback that trades
roughly five contamination points for one fewer hyperparameter (Qwen: 5.93\%
vs.\ 1.19\%).
\textbf{Recommended configuration: fixed $\tau{=}0.7$, $K{=}5$.}
The LogProb variant reaches near-perfect precision on \textsc{MemContam}
(1.000, 0.984, 0.999, 1.000 across the four backbones) at 23--28\,ms per fact,
a $12{-}14\times$ speedup over $K{=}5$, but this advantage is specific to
structured synthetic contexts; on \textsc{LoCoMo-Contam} soft scoring
outperforms LogProb on every backbone (\Cref{sec:analysis}), so we recommend
LogProb only for clean structured source data such as document or tool outputs.

\section{Failure Modes and Operating Regime}
\label{sec:analysis}

\paragraph{Operating regime.}
\method{} struggles in two situations.
When a candidate fact requires world knowledge beyond what the source context
provides, the LLM verifier may consistently vote ``yes'' for plausible-but-wrong
claims \citep{min2023factscore}; this is the structural limitation that
consistency is a proxy for correctness, not correctness itself
(\S\ref{sec:limitations}).
When relevant evidence is distributed across many prior conversation turns rather
than localized in a single retrievable chunk, the verifier cannot ground its
judgment and recall drops (Recall $=0.58$ on \textsc{LoCoMo-Contam} versus
$1.00$ on \textsc{MemContam}, \S\ref{sec:experiments}).
The gain is largest in the opposite regime: long trajectories where the cascade
has room to compound (\Cref{fig:cascade}), and high-recall low-precision writers
that leave aggressive extraction for the gate to filter.
\method{} is therefore best suited for document-grounded or tool-grounded memory
writes rather than for emergent multi-turn conversational memory.

\paragraph{Context structure moderates LogProb effectiveness.}
\label{sec:logprob_analysis}
The single-pass LogProb variant beats soft scoring on \textsc{MemContam} but loses
on \textsc{LoCoMo-Contam} on every backbone we tested (\Cref{tab:logprob_context}
in \Cref{app:logprob}). This realizes the bimodality asymmetry of
\S\ref{sec:logprob_method}: short structured contexts saturate the yes/no logits
so a single forward pass suffices, while long conversational contexts have
continuous score distributions that only $K$-sample averaging recovers. Use
LogProb on clean structured source data (document chunks, tool outputs,
knowledge-base entries) and soft scoring on open-ended dialogue.

\vspace{-1ex}
\section{Conclusion}
\label{sec:conclusion}
\vspace{-0.5ex}

We introduced \method{}, a write-time self-consistency gate that prevents
hallucinated facts from entering LLM agent memory.
The gate reduces contamination across two real-conversation benchmarks
(\textsc{LoCoMo-Contam}, \textsc{MSC-Contam}) and approaches a near-oracle upper
bound on a structured synthetic corpus (\textsc{MemContam}), with no fine-tuning;
contamination reductions on real data span 20\%--62\% across four LLM backbones,
with verifier capability emerging as the dominant factor in this spread.
The accompanying analysis isolates the \emph{contamination cascade}: write-time
errors compound across long trajectories, so admission decisions made early shape
memory quality much later in the interaction.
Correctness-based admission is therefore a necessary complement to existing
memory management, which has so far focused on utility and capacity.
We release code, the benchmarks, and the gate implementation.

\vspace{-1ex}
\section*{Limitations}
\label{sec:limitations}
\vspace{-0.5ex}

\method{} requires LLM access at write time and adds $O(K)$ cost per fact (the
LogProb variant uses one forward pass at 23--28\,ms, but only on short
structured contexts). Consistency is a proxy for correctness, so \method{}
cannot substitute for authoritative fact-checking. The main deployment risk is
implicit-fact recall: the gate rejects 42\% of correct probe facts on
\textsc{LoCoMo-Contam} (Recall $=0.58$, similar across all four backbones), a
cost pre-seeding masks here but that is permanent operationally
(see \S\ref{sec:analysis} for the operating regime in which the gate is
applicable).

\newpage
\bibliography{references}

\newpage
\appendix
\section*{Broader Impacts}
\label{sec:broader_impacts}

\method{} aims to make memory-augmented LLM agents more reliable by preventing
hallucinated facts from accumulating in long-term memory stores.
More reliable agent memory has broad positive applications: agents that assist with
medical information retrieval, legal research, or financial planning benefit
directly from reduced factual error propagation, and the gains are amplified in
exactly the long-horizon settings where these applications operate.
Because \method{} requires no fine-tuning and adds only a write-time check, it is
accessible to practitioners deploying existing models without specialized training
infrastructure.
We also release three benchmarks (\textsc{MemContam}, \textsc{LoCoMo-Contam},
\textsc{MSC-Contam}) so that the community can measure and compare memory
contamination methods.

Practitioners should be aware of two deployment-level risks.
First, passing a consistency check may create \emph{false assurance}: end users
may treat \method{}-filtered memory as factually verified when it has only been
checked for self-consistency against a potentially unreliable source context.
This risk is amplified in high-stakes domains (medical, legal, financial), where
practitioners must communicate clearly that consistency filtering is not a
substitute for authoritative fact-checking.
Second, at deployment scale, $K{=}5$ calls per fact increases energy consumption;
the LogProb variant ($\approx$27\,ms, single forward pass) substantially reduces
this overhead for latency- or energy-constrained deployments.
We do not foresee significant dual-use risks: write-time fact verification does not
introduce new capabilities for generating or spreading misinformation.

\clearpage
\section{Full Ablation Results}
\label{app:ablation}

\Cref{tab:ablation_full} extends \cref{tab:ablation} in the main paper with the complete
$K \in \{1,3,5,10\}$ and $\tau \in \{0.3,0.5,0.7,0.9,\text{adaptive}\}$ sweeps for all four
model backbones. Analysis and takeaways are in \cref{sec:ablation}.

\begin{table}[H]
\centering
\caption{Full ablation study on \textsc{MemContam} across all four backbones.
$K$-ablation rows use Adaptive $\tau$ (no-tuning fallback); $\tau$-ablation rows use $K{=}5$.
\textbf{Recommended} configuration: fixed $\tau{=}0.7$, $K{=}5$.
\textbf{Bold}: best per section per model.}
\label{tab:ablation_full}
\footnotesize
\setlength{\tabcolsep}{4pt}
\begin{tabular}{llcccc}
\toprule
& \textbf{Variant} & \textbf{Adm Prec} & \textbf{ContamRate} & \textbf{QA F1} & \textbf{Lat (ms)} \\
\midrule
\multicolumn{6}{@{}l}{\textit{Qwen2.5-32B-Instruct}} \\[0.3ex]
& \method{} ($K=1$)          & 0.931 & 6.89\% & 0.803 & \textbf{74.9} \\
& \method{} ($K=3$)          & 0.940 & 6.02\% & 0.807 & 205.7 \\
& \method{} ($K=5$, default) & 0.941 & 5.93\% & 0.803 & 341.9 \\
& \method{} ($K=10$)         & \textbf{0.944} & \textbf{5.57\%} & \textbf{0.814} & 682.7 \\
\cmidrule{2-6}
& Fixed $\tau = 0.3$         & 0.931 & 6.89\% & 0.801 & 344.0 \\
& Fixed $\tau = 0.5$         & 0.945 & 5.48\% & 0.809 & 347.3 \\
& Fixed $\tau = 0.7$         & \textbf{0.988} & \textbf{1.19\%} & \textbf{0.842} & 341.6 \\
& Fixed $\tau = 0.9$         & \textbf{0.988} & \textbf{1.19\%} & \textbf{0.842} & \textbf{340.4} \\
& Adaptive $\tau$ (default)  & 0.941 & 5.93\% & 0.803 & 341.9 \\
\cmidrule{2-6}
& LogProb                    & \textbf{1.000} & \textbf{0.00\%} & 0.824 & \textbf{27.5} \\
\midrule
\multicolumn{6}{@{}l}{\textit{Gemma4-31B-IT}} \\[0.3ex]
& \method{} ($K=1$)          & 0.964 & 3.57\% & 0.833 & \textbf{68.3} \\
& \method{} ($K=3$)          & 0.965 & 3.47\% & 0.833 & 203.8 \\
& \method{} ($K=5$, default) & \textbf{0.966} & \textbf{3.38\%} & 0.833 & 339.6 \\
& \method{} ($K=10$)         & 0.964 & 3.57\% & 0.833 & 679.0 \\
\cmidrule{2-6}
& Fixed $\tau = 0.3$         & 0.966 & 3.38\% & 0.833 & 339.7 \\
& Fixed $\tau = 0.5$         & 0.966 & 3.38\% & 0.833 & 339.7 \\
& Fixed $\tau = 0.7$         & \textbf{0.984} & \textbf{1.57\%} & \textbf{0.847} & 339.7 \\
& Fixed $\tau = 0.9$         & \textbf{0.984} & \textbf{1.57\%} & \textbf{0.847} & 339.7 \\
& Adaptive $\tau$ (default)  & 0.966 & 3.38\% & 0.833 & 339.7 \\
\cmidrule{2-6}
& LogProb                    & \textbf{0.984} & \textbf{1.57\%} & \textbf{0.847} & \textbf{24.3} \\
\midrule
\multicolumn{6}{@{}l}{\textit{Llama4-Scout-17B-16E}} \\[0.3ex]
& \method{} ($K=1$)          & 0.855 & 14.5\% & \textbf{0.717} & \textbf{54.3} \\
& \method{} ($K=3$)          & 0.859 & 14.1\% & 0.715 & 157.1 \\
& \method{} ($K=5$, default) & \textbf{0.861} & \textbf{13.9\%} & 0.716 & 264.6 \\
& \method{} ($K=10$)         & 0.860 & 14.0\% & 0.716 & 525.2 \\
\cmidrule{2-6}
& Fixed $\tau = 0.3$         & 0.857 & 14.3\% & 0.715 & \textbf{259.1} \\
& Fixed $\tau = 0.5$         & 0.860 & 14.0\% & 0.715 & 259.3 \\
& Fixed $\tau = 0.7$         & 0.997 & 0.30\% & 0.783 & 260.3 \\
& Fixed $\tau = 0.9$         & \textbf{0.999} & \textbf{0.10\%} & \textbf{0.785} & 259.6 \\
& Adaptive $\tau$ (default)  & 0.859 & 14.1\% & 0.718 & 259.7 \\
\cmidrule{2-6}
& LogProb                    & \textbf{0.999} & \textbf{0.10\%} & 0.784 & \textbf{23.1} \\
\midrule
\multicolumn{6}{@{}l}{\textit{Llama-3.3-70B-Instruct}} \\[0.3ex]
& \method{} ($K=1$)          & 0.998 & 0.20\% & 0.818 & \textbf{73.4} \\
& \method{} ($K=3$)          & 0.998 & 0.20\% & 0.818 & 218.1 \\
& \method{} ($K=5$, default) & 0.998 & 0.20\% & 0.818 & 363.4 \\
& \method{} ($K=10$)         & 0.998 & 0.20\% & 0.818 & 726.6 \\
\cmidrule{2-6}
& Fixed $\tau = 0.3$         & 0.998 & 0.20\% & \textbf{0.818} & 363.3 \\
& Fixed $\tau = 0.5$         & 0.998 & 0.20\% & \textbf{0.818} & 363.1 \\
& Fixed $\tau = 0.7$         & 0.998 & 0.20\% & 0.814 & 363.0 \\
& Fixed $\tau = 0.9$         & 0.998 & 0.20\% & 0.814 & 362.8 \\
& Adaptive $\tau$ (default)  & 0.998 & 0.20\% & \textbf{0.818} & \textbf{362.7} \\
\cmidrule{2-6}
& LogProb                    & \textbf{1.000} & \textbf{0.00\%} & \textbf{0.819} & \textbf{25.7} \\
\bottomrule
\end{tabular}
\end{table}

\section{Soft Scoring vs.\ LogProb by Context Structure}
\label{app:logprob}

\Cref{tab:logprob_context} reports the contamination-rate comparison referenced in
\S\ref{sec:logprob_analysis}.

\begin{table}[h]
\centering
\caption{Contamination rate $\rho$ (\%) of soft scoring ($K{=}5$, default) vs.\ LogProb on
\textsc{MemContam} (synthetic, structured) and \textsc{LoCoMo-Contam} (real conversations).
\textbf{Bold}: lower contamination per dataset per model.
LogProb dominates on \textsc{MemContam}; soft scoring dominates on \textsc{LoCoMo-Contam}.}
\label{tab:logprob_context}
\small
\setlength{\tabcolsep}{6pt}
\begin{tabular}{lcccc}
\toprule
& \multicolumn{2}{c}{\textsc{MemContam}} & \multicolumn{2}{c}{\textsc{LoCoMo-Contam}} \\
\cmidrule(lr){2-3}\cmidrule(lr){4-5}
\textbf{Model} & Soft $K{=}5$ & LogProb & Soft $K{=}5$ & LogProb \\
\midrule
Qwen2.5-32B-Instruct      & 5.93\% & \textbf{0.00\%} & \textbf{34.1\%} & 40.0\% \\
Gemma4-31B-IT             & 3.38\% & \textbf{1.57\%} & \textbf{18.9\%} & 23.7\% \\
Llama4-Scout-17B-16E      & 13.9\% & \textbf{0.10\%} & \textbf{29.7\%} & 34.3\% \\
Llama-3.3-70B-Instruct    & 0.20\% & \textbf{0.00\%} & \textbf{19.0\%} & 19.4\% \\
\bottomrule
\end{tabular}
\end{table}

\section{Prompt Templates}
\label{app:prompts}

{\sloppy
\paragraph{Benchmark generation prompt (used to construct \textsc{MemContam}).}
\begin{quote}
\texttt{You are creating a benchmark for testing memory quality in AI agents.\\
Given the following source context, generate:\\
1. Five (5) CORRECT atomic facts that are explicitly stated in the context.\\
2. Five (5) INCORRECT variants --- each based on a correct fact but subtly wrong\\
\phantom{2. }(swap an entity, change a number, negate, or introduce a plausible confabulation).\\
3. Three (3) question-answer pairs whose answers can be found in the correct facts.\\
Return a JSON object with keys: "correct\_facts", "incorrect\_facts", "questions".\\
Each question in "questions" is \{"question": "...", "answer": "..."\}.\\
Source context:\\
\{context\}\\
JSON output:}
\end{quote}

\paragraph{QA answering prompt (used to evaluate downstream QA F1).}
\begin{quote}
\texttt{You are a helpful assistant with access to a memory store.\\
Use the retrieved memories below to answer the question accurately.\\
Respond with a short, direct answer (1--5 words). Do not write full sentences.\\
Retrieved memories:\\
\{memories\}\\
Question: \{question\}\\
Answer (1--5 words):}
\end{quote}

\paragraph{Fact conversion prompt (LoCoMo-Contam: QA pair $\to$ declarative fact).}
\begin{quote}
\texttt{Given a question and its answer from a personal conversation, write a single,
self-contained factual statement that captures the key information.
The statement should read naturally, like a sentence from a diary or summary.\\
Question: \{question\}\\
Answer: \{answer\}\\
Output only the fact statement, nothing else.}
\end{quote}

\paragraph{Corruption prompt (LoCoMo-Contam and MSC-Contam: generate incorrect variant).}
\begin{quote}
\texttt{Below is a factual statement extracted from a personal conversation.
Your task: produce ONE corrupted version of this statement by changing exactly one
concrete detail (a name, date, place, number, or activity) to something plausible but WRONG.\\
Rules:\\
- Keep the sentence structure identical.\\
- Change only one detail.\\
- The corrupted version must directly contradict the original.\\
- Do NOT add hedging words like ``allegedly'' or ``reportedly''.\\
- Output only the corrupted statement, nothing else.\\
Original: \{fact\}\\
Corrupted:}
\end{quote}

\paragraph{Conversational turn prompt (LoCoMo-Contam and MSC-Contam: fact $\to$ natural utterance).}
\begin{quote}
\texttt{Convert the following fact into a short, natural first-person message that
\{speaker\} might send in a chat conversation.
The message should casually mention the fact as if referencing a shared memory.
Keep it to 1--2 sentences. Output only the message, nothing else.\\
Fact: \{fact\}\\
Message:}
\end{quote}

\paragraph{Memory extraction prompt.}
\begin{quote}
\texttt{Extract atomic, self-contained facts from the following context.\\
Each fact must be:\\
- A single complete sentence\\
- Directly stated in the context (no inferences)\\
- Specific enough to be verified independently\\
Return ONLY a JSON array of strings. No explanation.\\
Context:\\
\{context\}\\
Facts:}
\end{quote}

\paragraph{Consistency verification prompt (soft scoring, default).}
\begin{quote}
\texttt{You are a fact-checking assistant. Given a source context and a candidate fact,
rate how strongly the fact is supported by the context.\\
Source context:\\
\{context\}\\
Candidate fact:\\
\{fact\}\\
Question: How strongly is this candidate fact supported by the source context above?
Do not use outside knowledge.\\
0.0 = completely unsupported or directly contradicted by the context\\
0.5 = partially supported or ambiguous\\
1.0 = fully and directly supported by the context\\
Output only a single decimal number between 0.0 and 1.0.}
\end{quote}

\paragraph{Consistency verification prompt (binary, used for LogProb variant).}
\begin{quote}
\texttt{You are a fact-checking assistant. Given a source context and a candidate fact,
determine whether the fact is explicitly and directly supported by the context.\\
Source context:\\
\{context\}\\
Candidate fact:\\
\{fact\}\\
Question: Is this candidate fact directly and explicitly supported by the source context above?
Do not use outside knowledge. Answer with only "yes" or "no".}
\end{quote}
} 

\section{Qualitative Examples}
\label{app:qualitative}

\Cref{tab:qualitative} shows representative per-fact decisions made by \method{}
on \textsc{LoCoMo-Contam} (Qwen2.5-32B-Instruct, $K{=}5$, soft scoring).
Each row shows the extracted fact, the consistency score $\hat{p}$, the adaptive
threshold $\tau^*$, the admission decision, and the ground-truth label.

\begin{table}[h]
\centering
\caption{
    Representative \method{} admission decisions on \textsc{LoCoMo-Contam}.
    $\hat{p}$: consistency score (0--1). $\tau^*$: adaptive threshold.
    \cmark: correct decision; \xmark: error.
}
\label{tab:qualitative}
\small
\setlength{\tabcolsep}{4pt}
\begin{tabular}{p{7cm}ccccc}
\toprule
\textbf{Extracted fact} & $\hat{p}$ & $\tau^*$ & \textbf{Admitted} & \textbf{Label} & \textbf{Correct?} \\
\midrule
\multicolumn{6}{@{}l}{\textit{True Positives: correct facts correctly admitted}} \\[2pt]
Gina said dancing feels magical.
  & 1.00 & 0.80 & \cmark & correct & \cmark \\
Gina stressed how important it is to make their customers feel like they want to come back.
  & 1.00 & 0.80 & \cmark & correct & \cmark \\
Gina went to a dance class on July 21st.
  & 1.00 & 0.80 & \cmark & correct & \cmark \\
\midrule
\multicolumn{6}{@{}l}{\textit{True Negatives: corrupted facts correctly rejected}} \\[2pt]
Gina said \textit{painting} feels like magic.
  & 0.00 & 0.80 & \xmark & corrupted & \cmark \\
Gina visited Jon's \textit{old} studio.
  & 0.00 & 0.80 & \xmark & corrupted & \cmark \\
Jon and Gina both escaped into \textit{painting} last weekend.
  & 0.00 & 0.80 & \xmark & corrupted & \cmark \\
\midrule
\multicolumn{6}{@{}l}{\textit{False Positives: corrupted facts incorrectly admitted}} \\[2pt]
Jon and Gina attended a dance class together.
  & 1.00 & 0.80 & \cmark & corrupted & \xmark \\
Melanie is into pottery.
  & 1.00 & 0.80 & \cmark & corrupted & \xmark \\
\midrule
\multicolumn{6}{@{}l}{\textit{False Negatives: correct facts incorrectly rejected}} \\[2pt]
Calvin went to a car show with his dad.
  & 0.00 & 0.80 & \xmark & correct & \xmark \\
James mentioned a previous plan to meet up in Connecticut.
  & 0.00 & 0.80 & \xmark & correct & \xmark \\
\bottomrule
\end{tabular}
\end{table}

\paragraph{Analysis.}
True positives and true negatives show a clean separation: facts directly supported by
memory (e.g., ``dancing feels magical'') receive score 1.0, while facts that contradict
stored evidence (e.g., ``painting'' replacing ``dancing'') receive score 0.0.
False positives arise when the corrupted fact is \emph{consistent with but not identical to}
the correct fact in memory: ``Jon and Gina attended a dance class together'' is plausible
given Gina's dance class entry, even though the specific corruption was the date.
Similarly, ``Melanie is into pottery'' is supported by memory that lists pottery among
several of Melanie's hobbies, so the gate admits it despite the label counting it as corrupted.
False negatives occur when the correct fact is stated \emph{indirectly} or requires
multi-hop inference: ``Calvin went to a car show'' is correct but the memory stores a
longer compound sentence about Dave's engineering motivation, so the gate cannot directly
verify the sub-claim and scores it 0.0.
These failure modes motivate future work on multi-hop verification and
fine-grained corruption detection.

\section{Benchmark Dataset Details}
\label{app:benchmark}

\subsection{\textsc{MemContam}}
\label{app:memcontam}

\paragraph{Construction.}
\textsc{MemContam} contains 200 samples generated from five synthetic narrative contexts
(personal biography, library record, recipe, science fair report, spacecraft mission log).
Each context is repeated $200/5 = 40$ times with independent LLM-generated fact/corruption
sets, giving broad lexical variation despite sharing the same underlying entities.
For each sample, Qwen2.5-32B-Instruct is prompted to produce:
(i)~five \emph{correct} atomic facts explicitly stated in the context,
(ii)~five \emph{incorrect} variants---each derived from a correct fact by entity substitution,
number perturbation, negation insertion, or plausible confabulation,
and (iii)~three question-answer pairs answerable from the correct facts.
The dataset thus contains 1{,}000 correct facts, 1{,}000 incorrect facts, and 600 QA pairs.
By construction, presenting all facts to a WriteAll agent yields exactly $\rho = 50\%$
contamination.

\paragraph{Corruption taxonomy.}
Incorrect facts are generated via four strategies applied in priority order:
(1)~\textbf{Number perturbation}: numeric tokens are shifted by a small random delta
(e.g., ``34-year-old'' $\to$ ``35-year-old'');
(2)~\textbf{Negation}: the first copula is negated (``is'' $\to$ ``is not'');
(3)~\textbf{Proper-noun substitution}: a named entity is replaced with an anonymous
placeholder (``MIT'' $\to$ ``Another'');
(4)~\textbf{Contradicting clause}: a ``which is incorrect'' suffix is appended as a fallback.
This taxonomy is designed to produce corruptions that are \emph{semantically plausible}
but \emph{factually inconsistent} with the source context, the minimal condition for
stress-testing a consistency-based gate.

\paragraph{Splits and evaluation.}
All 200 samples are used as a single evaluation pool; there is no train split.
Experiments use three random seeds (42, 123, 456) that determine the order in which
facts are presented to the memory agent.
QA metrics (F1, EM) follow the SQuAD normalization: lowercased, punctuation-stripped
token overlap.

\paragraph{Source context templates.}
The five fixed contexts used to generate all 200 samples are reproduced verbatim below.

\begin{enumerate}[leftmargin=*, itemsep=4pt, topsep=4pt]
\item \textit{Personal biography.}
``Alice Smith is a 34-year-old software engineer at TechCorp who graduated from MIT in 2012.
She lives in Seattle with her husband David and their two children, Emma and Jack.
Alice enjoys hiking on weekends and has climbed Mount Rainier twice.
Her team at TechCorp is working on a new distributed database system called Pegasus.''

\item \textit{Library record.}
``The Westbrook Community Library was founded in 1923 and holds over 80,000 volumes.
The current director, Margaret Chen, has been in the role since 2018.
The library is open Monday through Saturday, 9am to 8pm, and Sunday 12pm to 5pm.
Annual membership costs \$25 for adults and \$10 for students.''

\item \textit{Recipe.}
``Recipe for Classic Tomato Basil Pasta: Boil 400g of spaghetti for 10 minutes.
In a separate pan, sauté 4 cloves of garlic in olive oil for 2 minutes.
Add 800g of crushed tomatoes and simmer for 15 minutes.
Stir in 30 fresh basil leaves and season with salt and pepper.
Serves 4 people. Total preparation time: 35 minutes.''

\item \textit{Science fair report.}
``The 2024 Greenfield Science Fair awarded first place to Sarah Liu, age 15,
for her project on photovoltaic efficiency improvements using perovskite coatings.
Second place went to Marcus Johnson for his water purification device.
The event was held at Lincoln High School on March 12, 2024, with 142 participants.''

\item \textit{Spacecraft mission log.}
``The Orion-7 spacecraft launched on June 3, 2023, from Cape Canaveral.
It carries a crew of three: Commander Elena Vasquez, Mission Specialist Dr.\ James Park,
and Flight Engineer Priya Nair. The mission duration is planned for 180 days.
Orion-7 will conduct experiments on bone density loss in microgravity.''
\end{enumerate}

\paragraph{License.}
\textsc{MemContam} is released under CC~BY~4.0.
The synthetic contexts are entirely model-generated; no copyrighted source text is included.

\subsection{\textsc{LoCoMo-Contam}}
\label{app:locomo_contam}

\paragraph{Source data.}
\textsc{LoCoMo-Contam} is derived from LoCoMo-10 \citep{maharana2024locomo}, a set of 10
long-term dyadic conversations (400--700 turns each, 19--32 sessions per conversation).
We use all 10 conversations without filtering.

\paragraph{Probe construction.}
For each conversation, we select up to $N{=}5$ QA pairs from LoCoMo's annotations,
restricting to factual and temporal categories with concrete, verifiable answers.
Each QA answer is converted to a self-contained declarative fact using Qwen2.5-32B-Instruct
(\eg ``Caroline moved from Sweden 4 years ago'').
A corrupted variant is generated by prompting the model to change exactly one concrete
detail (name, number, date, or place) to a plausible but incorrect alternative.
Both the correct and corrupted facts are paraphrased into natural conversational utterances
as if spoken by one of the conversation participants.
The resulting benchmark has 50 probe pairs (5 per conversation) at 50\% contamination.

\paragraph{Evaluation protocol.}
The agent first processes the full conversation to populate memory.
The correct probe facts are then pre-seeded directly into memory (bypassing the gate) to
ensure retrieval evidence is always available.
The gate is evaluated by presenting each probe turn through the full observe pipeline:
the memory writer extracts a fact, the gate queries memory for supporting context, and
an admission decision is recorded.

\paragraph{License.}
LoCoMo is released under CC~BY~NC~4.0; our use is non-commercial research only.

\subsection{\textsc{MSC-Contam}}
\label{app:msc_contam}

\paragraph{Source data.}
\textsc{MSC-Contam} is derived from the Multi-Session Chat (MSC) dataset
\citep{xu2022msc}, comprising 20 dyadic conversations (2 sessions each),
grounded in evolving speaker personas.
We use all 20 conversations.

\paragraph{Probe construction.}
MSC conversations do not provide pre-annotated QA pairs; instead, speakers are given
persona sentences at session start.
We filter persona sentences for concreteness, keeping only those with a verifiable
concrete detail (name, number, place, or occupation), and select up to $N{=}5$ persona
facts per conversation.
Each fact is corrupted by changing exactly one concrete detail using the same prompts as
\textsc{LoCoMo-Contam}, then paraphrased into a natural conversational utterance.
QA pairs for downstream evaluation are generated directly from the original persona facts.
The resulting benchmark has 41 probe pairs (up to 5 per conversation $\times$ 20 conversations;
conversations with fewer than 5 concrete persona facts contribute fewer probes)
at 50\% contamination.

\paragraph{Evaluation protocol.}
Same as \textsc{LoCoMo-Contam}: all session turns are processed to populate memory,
correct facts are pre-seeded, and each probe is evaluated through the full gate pipeline.

\paragraph{License.}
MSC is released under the MIT License.


\end{document}